\documentclass[letterpaper, 10 pt, conference]{ieeeconf}  

\IEEEoverridecommandlockouts                              

\usepackage{graphics} 
\usepackage{epsfig} 
\usepackage{amsmath} 
\usepackage{amssymb}  

\usepackage{xcolor}
\usepackage{cite}
\usepackage{mathtools}  
\usepackage{algorithmic}
\usepackage{algorithm}
\usepackage{booktabs}
\usepackage[bottom]{footmisc}
\usepackage{multirow}
\usepackage{hyperref}
\usepackage{adjustbox}
\usepackage{capt-of}
\title{\LARGE \bf Online Safety Property Collection and Refinement\\for Safe Deep Reinforcement Learning in Mapless Navigation
}

\author{Luca Marzari$^{1,\dagger,*}$, Enrico Marchesini$^{2,\dagger}$ and Alessandro Farinelli$^{1}$
\thanks{$^{1}$ Department of Computer Science, University of Verona, Italy.}%
\thanks{$^{2}$ Khoury College of Computer Sciences, Northeastern University, US.}
\thanks{$^{\dagger}$ Equal contribution}
\thanks{$^{*}$ Contact author: {\tt\small luca.marzari@univr.it}}%
}

\begin{document}

\maketitle
\thispagestyle{empty}
\pagestyle{empty}


\begin{abstract}
Safety is essential for deploying Deep Reinforcement Learning (DRL) algorithms in real-world scenarios.
Recently, verification approaches have been proposed to allow quantifying the number of violations of a DRL policy over input-output relationships, called properties. However, such properties are hard-coded and require task-level knowledge, making their application intractable in challenging safety-critical tasks. To this end, we introduce the Collection and Refinement of Online Properties (CROP) framework to design properties at training time. CROP employs a cost signal to identify unsafe interactions and use them to shape safety properties. Hence, we propose a refinement strategy to combine properties that model similar unsafe interactions.
Our evaluation compares the benefits of computing the number of violations using standard hard-coded properties and the ones generated with CROP. We evaluate our approach in several robotic mapless navigation tasks and demonstrate that the violation metric computed with CROP allows higher returns and lower violations over previous Safe DRL approaches. 
\end{abstract}
\section{Introduction}
\label{sec:introduction}
Recently, Deep Reinforcement Learning (DRL) algorithms achieved significant results in robotic applications, ranging from manipulation\cite{manipulation1, manipulation2} to mapless navigation\cite{icra_marl, drl_navigation3, iros_marl}. However, applying these techniques in real-world scenarios is seldom straightforward as non-linear function approximators are vulnerable to adversarial inputs\cite{adversarial, amir2022verifying}. Given such issues, it is crucial to employ verification techniques and safety metrics in a safety-critical context.

To this end, Safe DRL techniques\cite{Garcia2015} have been investigated to enhance safety in robotic tasks. In particular, Safe DRL problems are typically modeled using Constrained Markov Decision Processes (CMPDs)\cite{ConstrainedMDP}, where an agent aims at maximizing a reward signal while keeping cost values accumulated upon visiting unsafe states under a hard-coded threshold. However, the constraints imposed by these approaches hinder exploration, failing to learn safe behaviors in complex environments \cite{SafetyGym, AAAI_Sos}. Alternative ways have been investigated to overcome the difficulty of designing Safe DRL algorithms that combine the concept of risk in the optimization while avoiding unsafe situations\cite{Garcia2015, gecco, Verification}. In particular, recent methods rely on Formal Verification (FV) \cite{Verification} to quantify the number of correct decisions a policy chooses over desired safe specifications and use such information to evaluate the agents' decision-making. In more detail, a verification framework checks hard-coded input-output relationships (i.e., safety properties) in a domain of interest, verifying the decision-making process of a DRL policy. Hence, given a particular configuration of the agent's state space, i.e., a space $x \subset S$, a verification process identifies all the states where the policy selects the action $y \in A$ (where $A$ is the agent's action space) that violates the safety properties.
However, FV is an \textit{NP-Complete} problem \cite{reluplex}, and it is thus unfeasible to use such violation information to foster safety during the training. 

To this end, \cite{AAAI_Sos, aamas_safety} proposed a sample-based approximation method to enumerate the number of states in the space $x$ that violate a specific property. Such a value referred to as \textit{violation}, has been used to induce safety information during the training. The \textit{violation} is particularly beneficial in a safety-critical setup, as it includes safety information based on the hard-coded specified properties, thus, allowing foster desired safe behaviors in the agent.

 \begin{figure}[t]	
    \begin{center}
    	\includegraphics[width=\linewidth]{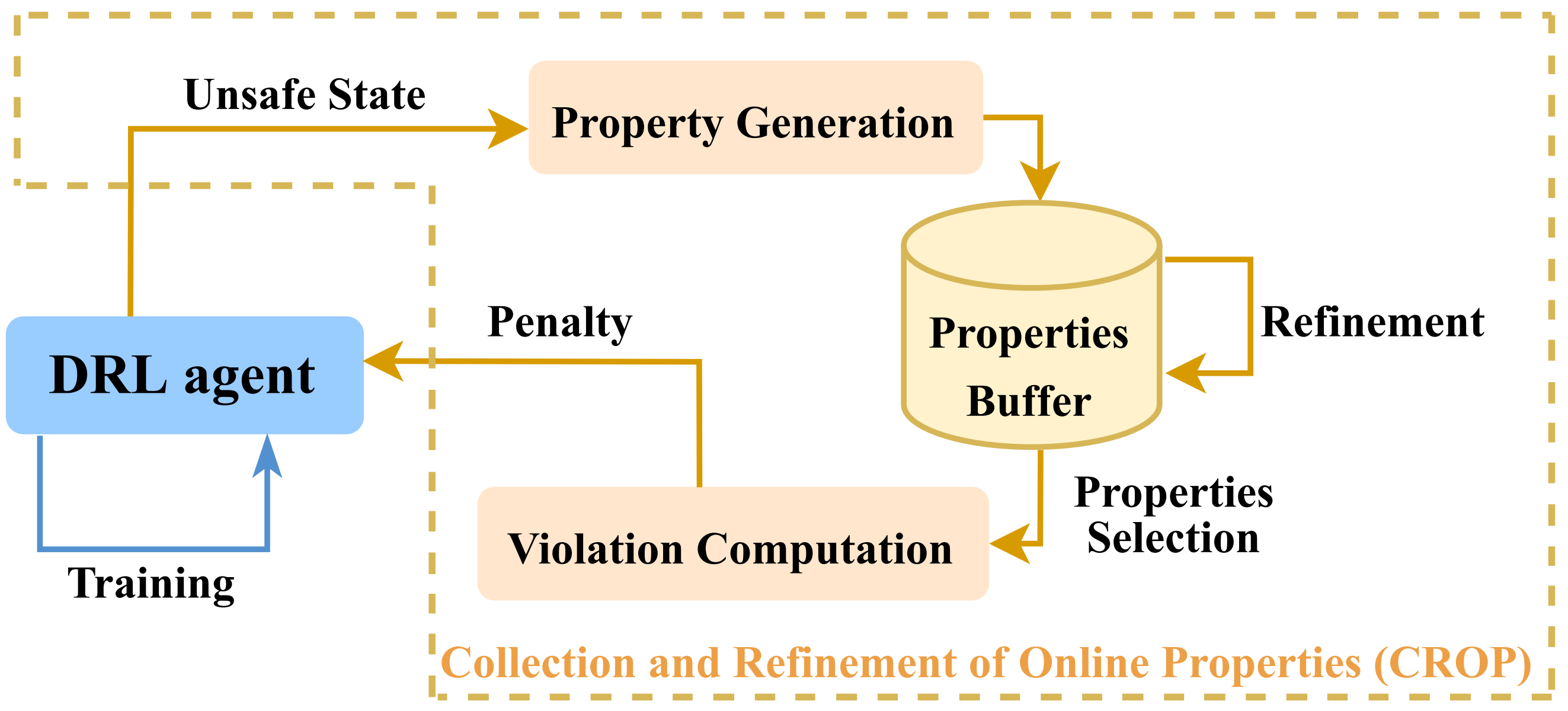}
        \caption{Overview of the proposed CROP framework.}
        \label{fig:overall}
    \end{center}
    \vspace{-5mm}
\end{figure}
    
Modeling hard-coded properties, however, is a severe limitation of verification approaches for two main reasons. First, the properties assume having precise task-level knowledge regarding the robotic task (e.g., presence of fixed or dynamic obstacles, their shape, etc.) and agent specification (e.g., in the case of a mobile robot, the agent's dimension, linear and angular velocity $\Vec{v}$). Second, writing an exhaustive set of properties to cover all the unsafe behaviors is impractical in dynamic scenarios. 
These limitations significantly impact applying the \textit{violation} in complex safety-critical domains where writing safety properties is not trivial. For example, it could be unfeasible to write properties for robots with complex dynamics (e.g., quadrupeds with multiple joints), as shown in \cite{AAAI_Sos}. For this reason, prior works that consider the violation in DRL contexts \cite{iros_aquatic, curriculum, pore2021safe} use a limited set of 3-5 high-level properties to provide the agent with core safe behaviors specification (e.g., in the case of robotic navigation, do not turn right if close to an obstacle). Hence, the agent may experience unsafe interactions not encoded by the manually specified set of safety properties. 

Despite the limitations, previous violation-based approaches \cite{Liu_violation} achieved comparable or better results over commonly employed Safe DRL approaches such as Constrained DRL\cite{ConstrainedDRL}. Motivated by this, we address the issue of hard-coding properties by providing a method that can collect and refine safety specifications during training.

Specifically, we propose the Collection and Refinement of Online Properties (CROP) framework (Fig. \ref{fig:overall}) to collect and refine safety properties during training time. Similarly to prior Safe DRL literature \cite{IPO, Lagrangian, LagrangianBehavior}, CROP uses an indicator function, called cost, that deems a state-action interaction as unsafe. When the agent performs an unsafe interaction, CROP generates a safety property using the state and the action that led to the hazardous situation. Hence, CROP performs a \textit{refinement} procedure to combine similar properties collected online. In particular, CROP uses a similarity rule to compute whether two safety properties encode similar unsafe interactions. Hence, our method merges similar properties into one that incorporates both the unsafe states. Finally, the violation is computed using a previous sample-based approximation method \cite{AAAI_Sos} on CROP's dynamic set of online generated properties and is used as a penalty for the training.

Following recent works on Safe DRL\cite{SafetyGym, AAAI_Sos, drl_navigation1, drl_navigation2}, we focus our evaluation on different mapless navigation domains and compare our method with (i) a penalty-based method that uses hard-coded safety properties, and (ii) a Constrained DRL algorithm. Our evaluation relies on realistic Unity environments\cite{unity} that enable the transfer of policies trained in simulation on ROS-enabled platforms. We provide a video (\url{shorturl.at/drVW6}) with the real-world experiments of the policies trained in our environments using the \href{https://www.turtlebot.com/turtlebot3/}{TurtleBot3} as an agent. The empirical evaluation confirms that the CROP framework addresses the lack of information provided by the hard-coded properties approaches. Crucially, our results show that using online generated properties lead to a more robust \textit{violation} computation that translates into safer behaviors (i.e., reduced number of collision during the training and evaluation in a previously unseen scenario). Moreover, we provide formal guarantees on the policy behaviors by employing a formal verification analysis of the trained agents \cite{prove}. 

This work makes the following contributions to the state-of-the-art: (i) we propose CROP, a novel framework for collecting properties online, overcoming the limitations of hard-coding such properties. (ii) We show how to refine similar safety properties to limit the number of generated properties and improve the design of the input-output relationships. (iii) We provide a quantitative empirical evaluation with prior approaches based on hand-designed properties and a standard Constraint DRL method. Moreover, we perform a qualitative validation on the real robot. 
\section{Preliminaries and Related Work}
\label{sec:preliminaries}

\subsection{Safety property}\label{safety_prop}
Safety properties are input-output relationships. In more detail, given a Deep Neural Network (DNN) $\mathcal{F}$ with $y_{1,\dots, n}$ possible outputs and a subspace of the state space $x \subset S$, a property is defined as \cite{Verification}:
\begin{equation}
    \mathcal{P}: \underbrace{x \subset S}_{\mathcal{P}[x]} \implies \underbrace{y_k < y_i}_{\mathcal{P}[y]} \quad\forall i \in [1, n]-\{k\}
    \label{eq:safety_prop}
\end{equation}

\begin{figure}[t]	
    \begin{center}
        \includegraphics[width=0.9\linewidth]{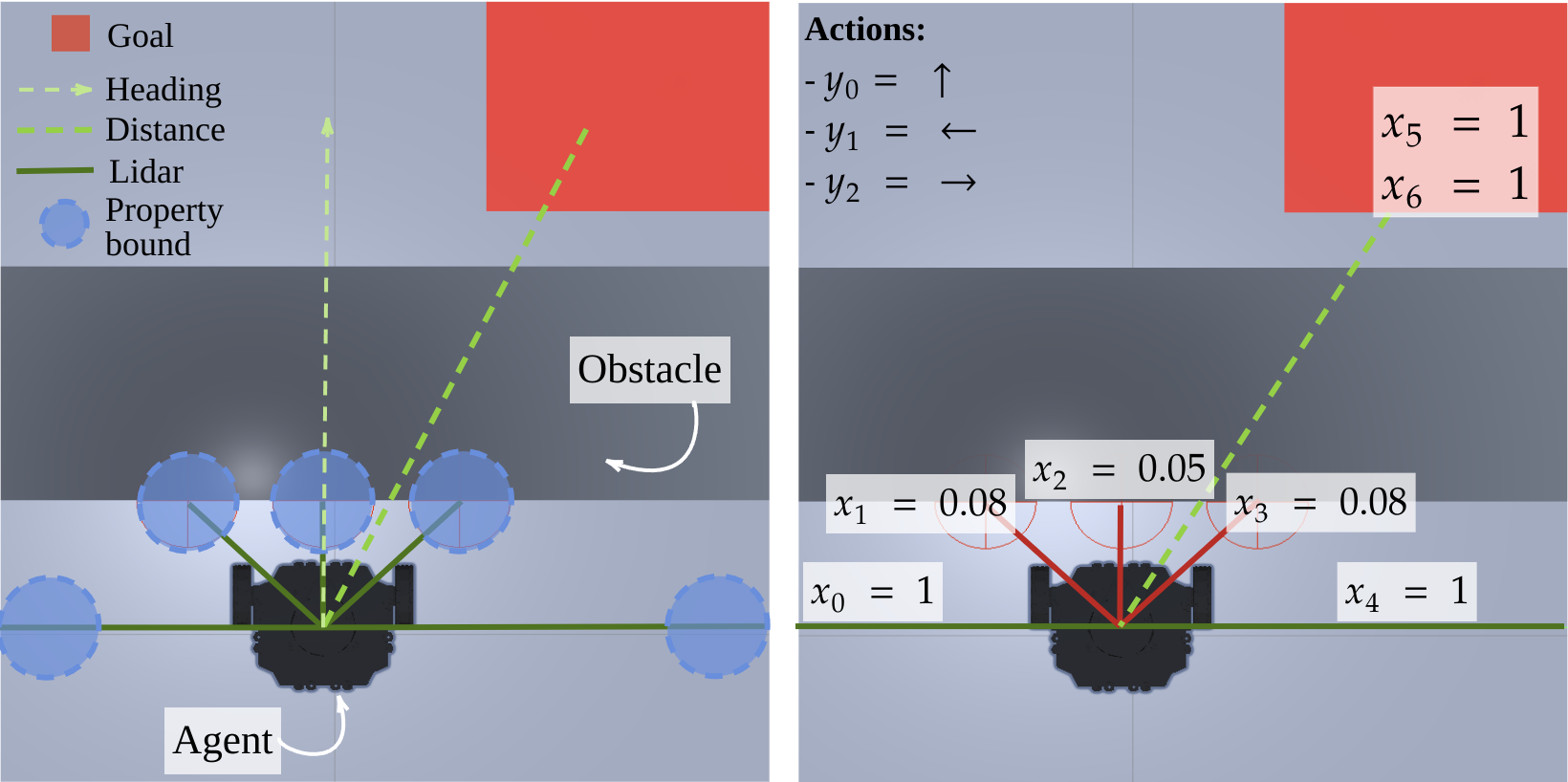}
        \caption{Left: components of a safety property. Right: Explanatory image of a safety property $\mathcal{P}_\uparrow$ for a mapless navigation context.}
        \label{fig:property}
    \end{center}
    \vspace{-8mm}
    \end{figure}

Typically, the subspace $x$ is encoded as a property $\mathcal{P}$ (i.e., $\mathcal{P}[x]$) using intervals to shape the surroundings of an unsafe state of interest. Hence, $\mathcal{P}[x]$ is a vector of intervals, one for each value of the observation space $S$. Given our interest in verifying DRL policies, the inequality of Eq. \ref{eq:safety_prop} $\mathcal{P}[y]$, encodes a condition as \textit{never select the action corresponding to the output $y_k$}. For instance, in a value-based framework where the policy chooses the action with the highest value, the post-condition is implemented to verify that the selected action does not correspond to the action that would violate the safety property. A similar check can be achieved in a policy gradient setup by using deterministic policies for evaluation and incorporating stochastic policies for exploration. 
For example, prior works \cite{superl, curriculum} in mapless navigation tasks use the agent's maximum velocity $\Vec{v}$ to compute the minimum distance that can be reached before colliding with an obstacle. Fig.\ref{fig:property} (right) shows a practical example where, knowing that a distance greater than $\pmb{\epsilon}=0.05$ prevents the agent from colliding at maximum speed, a safety property checks that an agent does not move forward while there is an obstacle in front. Following Eq. 1, this is formally encoded as:
\begin{align*}\label{eq:safety_prop}
    \mathcal{P}_{\uparrow}&:  x_0,x_4 \in [0.95, 1] \wedge x_1,x_3\in [0.03, 0.08]\wedge \\ 
    &\underbrace{x_2\in [0, 0.05] \wedge x_{5},x_{6}\in [-1, 1]}_{\mathcal{P}_{\uparrow}[x]} \implies \underbrace{y_0 < y_1, y_2}_{\mathcal{P}_{\uparrow}[y]} \quad
\end{align*}

where $x_0, \dots, x_{4}$ are the 5 lidar values, $x_{5}, x_{6}$ is the relative position of the goal with respect to the agent's position (i.e., distance and heading), and action $y_0$ corresponds to a forward movement. In the example, we set $x_5, x_6$ equal to $[-1, 1]$ to indicate a generic goal position in the environment. Broadly speaking, if the agent is in a state described by $\mathcal{P}_{\uparrow}[x]$, it should not choose action $y_0$ that corresponds to a forward movement (hence, leading to a collision). 

\subsection{Sample-based violation computation}\label{verification}
Due to the computational demands of the verification frameworks\cite{Verification}, and the \textit{NP-Completeness} of DNN Formal Verification (FV)\cite{reluplex}, computing the \textit{violation} value in the training loop is unfeasible. To this end, we compute an approximate \textit{violation} using a sample-based method to enumerate the number of states in the subset $x$ that violate a property $\mathcal{P}$, as proposed by \cite{aamas_safety, CountingProVe}. In detail, these approaches require $\mathcal{F}$ and a set $\mathbb{B}$ of safety properties that contain, in the subspace $x$, a surrounding of an unsafe state. For each property's subspace $\mathcal{P}[x] \in \mathbb{B}$, we sample a set of \textit{m} random points (i.e., states). These points represent a set of possible unsafe states the agent could reach in that subspace $\mathcal{P}[x]$. Finally, we perform a DNN forward propagation of these unsafe states and quantify the ratio of unsafe decisions of a policy $\pi$ over the number of sampled points.

\subsection{Penalty-based objective for safety}
We model the navigation tasks as a Markov Decision Process (MDP), described by a tuple $<S,A,R,P,\gamma>$ where $S$ is the observation space, $A$ is the action space, $P : S \times A \to S$ is the
transition function and $R : S \times A \to \mathbb{R}$ is the reward function. $\gamma \in [0, 1]$ is the discount factor that allows the control of the influence of future rewards. Given a stationary policy $\pi \in \prod$, the agent aims to maximize $\mathbb{E}_{\tau \sim \pi} \left[\sum_{t=0}^\infty \gamma^t R(s_t, a_t)\right]$, i.e., the expected discounted return for each trajectory $\tau = (s_0, a_0, \cdots)$.

Recent works \cite{Garcia2015, reward_shaping1, reward_shaping2} show how incorporating penalties into the reward effectively influences policy toward safer behaviors while avoiding the limitations introduced by constrained approaches \cite{Lagrangian}.
Following this direction, we incorporate the \textit{violation} value in a penalty-based objective:
\begin{equation}
    \max_{\pi \in \Pi} J_{R}^{\pi} :=  \mathbb{E}_{\tau \sim \pi} \left[\sum_{t=0}^\infty \gamma^t R(s_t, a_t) - Z(\cdot)\right]
\label{eq:penaltyobj}
\end{equation}
where $Z(\cdot)$ is a penalty function (the cost or the \textit{violation}). 


\section{Collection and Refinement \\of Online Properties}
\label{sec:method}
The general flow of CROP is presented in Algorithm \ref{alg:CROP}. In particular, we augment the DRL training with a buffer $\mathbb{P}$ that stores all the properties generated by CROP (line 3). In more detail, given a state $s_t$ deemed unsafe by the cost signal, CROP generates a new safety property $\mathcal{P}'$ considering an $\pmb{\epsilon}$ surrounding of the state $s_{t-1}$,  and the action $a_{t-1}$ that triggered a positive cost (i.e., considering the state-action interaction as unsafe) (line 5). In particular, CROP models the subspace of $\mathcal{P}'[x]$ with a set of intervals written as $[x_i-\; \pmb{\epsilon},\; x_i] \quad \forall\; x_i \in s_{t-1}$. The implication $\mathcal{P}'[y]$ checks that the agent does not choose the action $a_{t-1}$ that led to the unsafe state $s_t$. Note that the $\pmb{\epsilon}$ required by CROP differs from the epsilon value required to design a hard-coded safety property (as in the example of Sec.\ref{safety_prop}), which requires task-level knowledge. Specifically, in CROP, $\pmb{\epsilon}$ is initialized as an arbitrarily small value that will be shaped during refinement. 
Subsequently, CROP selects the set of properties that contain the state $s_{t-1}$ in their subspaces $\mathcal{P}[x]$ and that encode the implication $\mathcal{P}[y]$ over the same action $a_{t-1}$ (line 6). Hence, for each property selected, our method verifies whether there is a similar property $\mathcal{P}$ with respect to $\mathcal{P}'$ that needs to be refined (lines 7-14). In particular, CROP uses the refinement rule described in the next section to consider the similarity between properties.
\begin{algorithm}[t]
    \small
    \caption{Collection and Refinement of Online Properties}
    \label{alg:CROP}

    \begin{algorithmic}[1]
    \STATE \textbf{Given: } 
    \begin{itemize}
        \item a DRL agent parametrized by a DNN $\mathcal{F}_\theta$
        \item $\pmb{\epsilon}$ an initial size for a surrounding of an unsafe state to consider
        \item $\sim$ rule of similarity for safety properties as in Sec.\ref{prop_sim}.
    \end{itemize}
    
    \STATE During each episode of the training of DRL agent:
    \STATE $\mathbb{P} \gets \emptyset$
    \IF{$s_{t}$ is unsafe (i.e., cost $> 0$)}
        \STATE $\mathcal{P'} \gets$ \textit{Generate Property($\pmb{\epsilon}$, $s_{t-1}, a_{t-1}$)} \hspace*{\fill} $\rhd$ as Sec.\ref{sec:method}
        

        \FOR{$\mathcal{P} \in \mathbb{P}$ where  $s_{t-1} \subseteq \mathcal{P}[x] \wedge a_{t-1}  = y_k \in \mathcal{P}[y]$}
            \IF{$ \not \exists\; \mathcal{P} \sim \mathcal{P'}$} 
                \STATE $\mathbb{P} \gets \mathbb{P} \cup \mathcal{P'}$
            \ELSE
                \STATE $\mathbb{P}\gets$ \textit{Property Refinement}($\mathcal{P},\mathcal{P}',\mathbb{P}$) \hspace*{\fill} $\rhd$ as Sec.\ref{prop_sim}
            \ENDIF
            
        \ENDFOR

    \STATE violation $\gets$ \textit{violation computation}($\mathcal{F}_\theta$, $\mathbb{P}$) \hspace*{\fill} $\rhd$ as in Sec.\ref{verification}
    \ENDIF
\end{algorithmic}
\end{algorithm}
However, the collection of properties following this procedure remains challenging in domains where many unsafe interactions occur (i.e., the buffer $\mathbb{P}$ grows significantly). Hence, we decide to reset the properties buffer at the beginning of each episode. This approach allows us to compute the \textit{violation} to be given as a penalty only on the unsafe situations encountered during the trajectories seen during the episode. Finally, we compute the \textit{violation} using the sample-based estimation discussed in Section \ref{verification} (line 16).

\subsection{Properties similarity and refinement process}\label{prop_sim}
\begin{figure}[b]
    \vspace{-3mm}
    \begin{center}
    	\includegraphics[width=\linewidth]{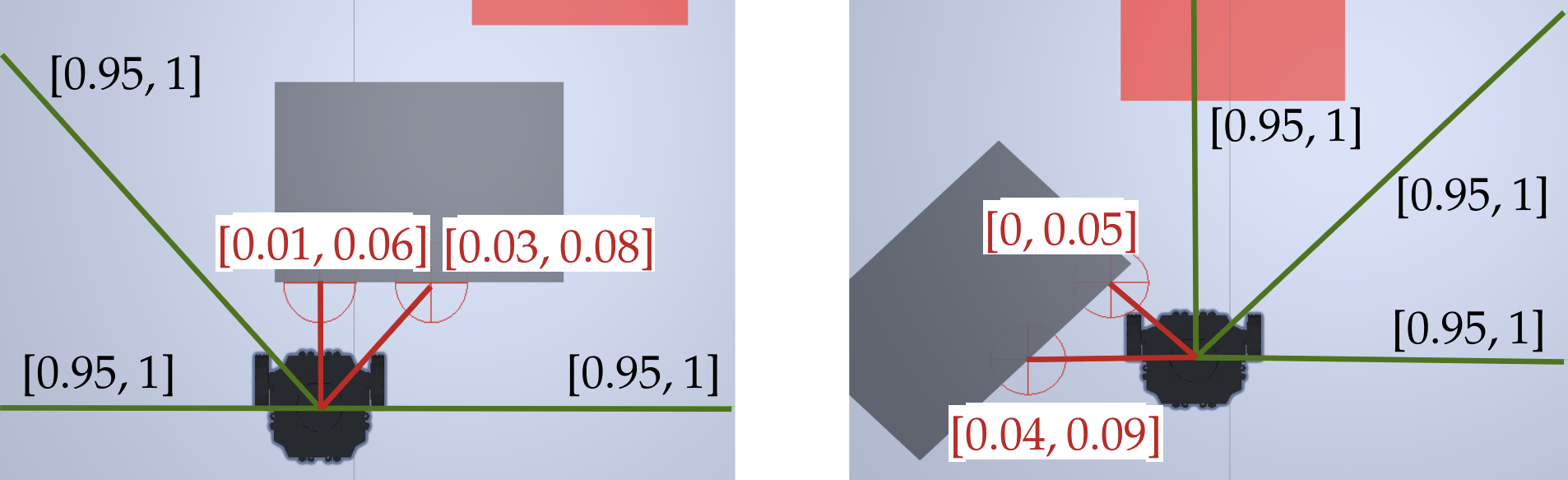}
        \caption{Explanatory example of two different safety properties for the same action $y_k$ that encodes a forward movement.}
        \label{fig:diff_props}
    \end{center}
    \vspace{-7mm}
\end{figure}

Our method requires a rule to define when two safety properties are \textit{not similar}. To this end, we illustrate a possible scenario in the context of mapless navigation in Fig. \ref{fig:diff_props}. 
In particular, given the nature of a safety property, it is conceivable to find two or more different safety properties for the same action $y_k$ of a DNN. As an example, in both the scenarios depicted in Fig. \ref{fig:diff_props}, the agent would collide upon moving forward. However, the subspaces of the safety properties $\mathcal{P}[x]$ significantly differ (for simplicity, we omit the values for heading and distance from goal). Hence, we detect \textit{not similar} properties leveraging Moore's interval algebra\cite{moore}. In particular, we define this rule:
\begin{align*}
    \mathcal{P}  \not\sim \mathcal{P'} \iff &\exists\; i \in [0, |x|] \;s. t \\
    &\big(\mathcal{P}[x]_i \subseteq \psi \vee \mathcal{P}'[x]_i \subseteq \psi \big) \wedge\\ 
    & |\mathcal{P}[x]_i - \mathcal{P'}[x]_i| > \beta 
\end{align*}
Broadly speaking, given two safety properties $\mathcal{P}$ and $\mathcal{P'}$, the above rule investigates whether there is at least one index $i\in [0, |x|]$ of the properties subspaces $\mathcal{P}[x]_i$ or $\mathcal{P}'[x]_i$, which is contained in $\psi$ (i.e., an interval that encodes a potentially unsafe situation. For example, in Fig. \ref{fig:diff_props} on the right, $\psi$ equals [0, 0.09] and is highlighted in red). Moreover, CROP checks whether the lower bound of the absolute difference in value between the two considered intervals $|\mathcal{P}[x]_i - \mathcal{P}'[x]_i|$ is greater than a threshold value $\beta$. If these two conditions are true for at least one index $i$ in the properties, then we consider $\mathcal{P}$ and $\mathcal{P}'$ \textit{not similar}. The value $\beta$ can be tuned to be more or less restrictive about considering two properties \textit{not similar}. Our similarity rule successfully distinguishes two different properties in the situations depicted in Fig. \ref{fig:diff_props}. For example, for the first lidar scan on the left, we have at least one of the two intervals in $[0, 0.09]$ and the lower bound of the absolute difference between the two intervals is equal to $0.86$ and, therefore, above a threshold $\beta = 0.1$. Crucially, similar considerations are applicable to different tasks to apply our collection and refinement procedure.

\begin{figure}[h!]
    \vspace{-3mm}
    \begin{center}
    	\includegraphics[width=\linewidth]{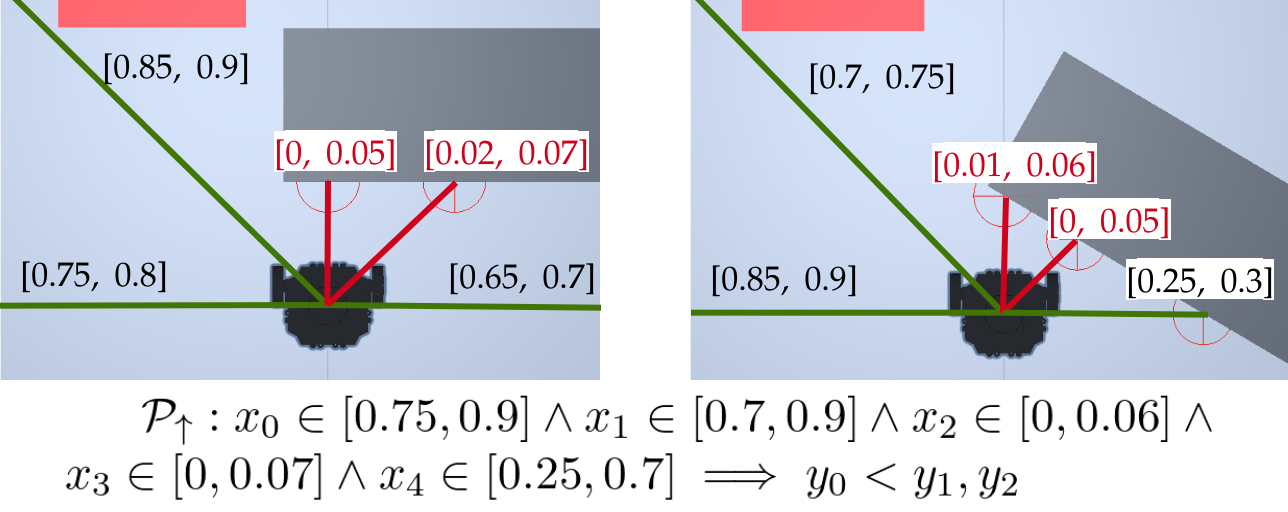}
        \caption{Explanatory example of refinement process for two \textit{similar} safety properties.}
        \label{fig:sim_props}
    \end{center}
    \vspace{-7mm}
\end{figure}

On the other hand, if two properties are \textit{similar}, the refinement process is performed, merging the two properties into one, considering the minimum lower and maximum upper bound for each interval. 
An explanatory image is reported in Fig. \ref{fig:sim_props}.

\section{Experiments}
\label{sec:evaluation}
Our evaluation focuses on a mapless navigation task, a well-known benchmark in the recent Safe DRL literature\cite{SafetyGym, AAAI_Sos, drl_navigation1, drl_navigation2}. In particular, in this task, a robot has to learn to navigate towards a random target using only local observation and the target position, without a map of the surrounding environment or obstacles. Similar to prior works \cite{drl_navigation2, curriculum}, we consider an observation space with 21 sparse laser scans sampled at a fixed distance in a 360 degrees range and two values for the target position relative to the robot (i.e., distance and heading) as observation space. In addition, to reduce training times while maintaining good navigation skills, we use a discrete action space\cite{drl_navigation2} that encodes angular and linear velocities. In our experiment, we use six discrete actions to encode the following movements: a forward movement, a rotation with and without linear velocity, and a stationary position. This allows us to capture the different types of movements accurately. 

We implement CROP in a policy-based Proximal Policy Optimization (PPO) \cite{PPO} baseline using a penalty shaping as in Eq. \ref{eq:penaltyobj}. We compare our method with previous violation-based penalty approaches that use hand-designed properties (i.e., PPO\_violation), a cost-based penalty (i.e., PPO\_cost), and a standard constraint approach, i.e., Lagrangian PPO (LPPO)\cite{Lagrangian}.
We use standard hyper-parameters from earlier research\cite{PPO, SafetyGym, AAAI_Sos} consisting of two hidden layers with 64 units each and activated with ReLU to model actors and critic networks, respectively. Moreover, to ensure a fair comparison with a standard constraint approach (LPPO), we set the cost threshold to the average cost produced by the penalty-based baselines. All the evaluated approaches use the same base reward function consisting of a dense reward determined by the difference in distance ($\Delta$) between the target and the agent's position. 
Regarding previous PPO\_violation approach, we consider a set of hard-coded proprieties to test basic safety behaviors, such as \textit{do not perform a \{forward, left, right\} movement, if there is an obstacle near to the \{back, front, left right\}}. Specifically, we consider the same properties of prior work \cite{prove} plus the properties to cover the back of the robot due to dynamic obstacles.
Moreover, for PPO\_CROP we set $\beta$ equal to $0.1$ as it achieves a good trade-off between the number of properties collected and the safety information provided to the agent. We also tuned a cloud size of $m=10.000$ points and we reset the buffer of properties at each episode as in Alg.\ref{alg:CROP} for our CROP method.
Data are collected on a RTX 2070, and an i7-9700k, reporting the mean and standard deviation gathered from ten independent runs.

Our empirical evaluation aims at investigating whether the CROP-based approach can increase return and reduce the number of violations. Crucially, a lower number of violations translate into fewer collisions (i.e., the agent chooses fewer actions that potentially lead to a collision). Hence, a lower violation value directly maps into a lower cost.
 
\subsection{Environments}
We use two training environments to evaluate our methodologies, called \textit{Fixed obstacles} and \textit{Dynamic obstacles}, respectively. 
\begin{figure}[h!]	
    \begin{center}
    	\includegraphics[width=0.5\linewidth]{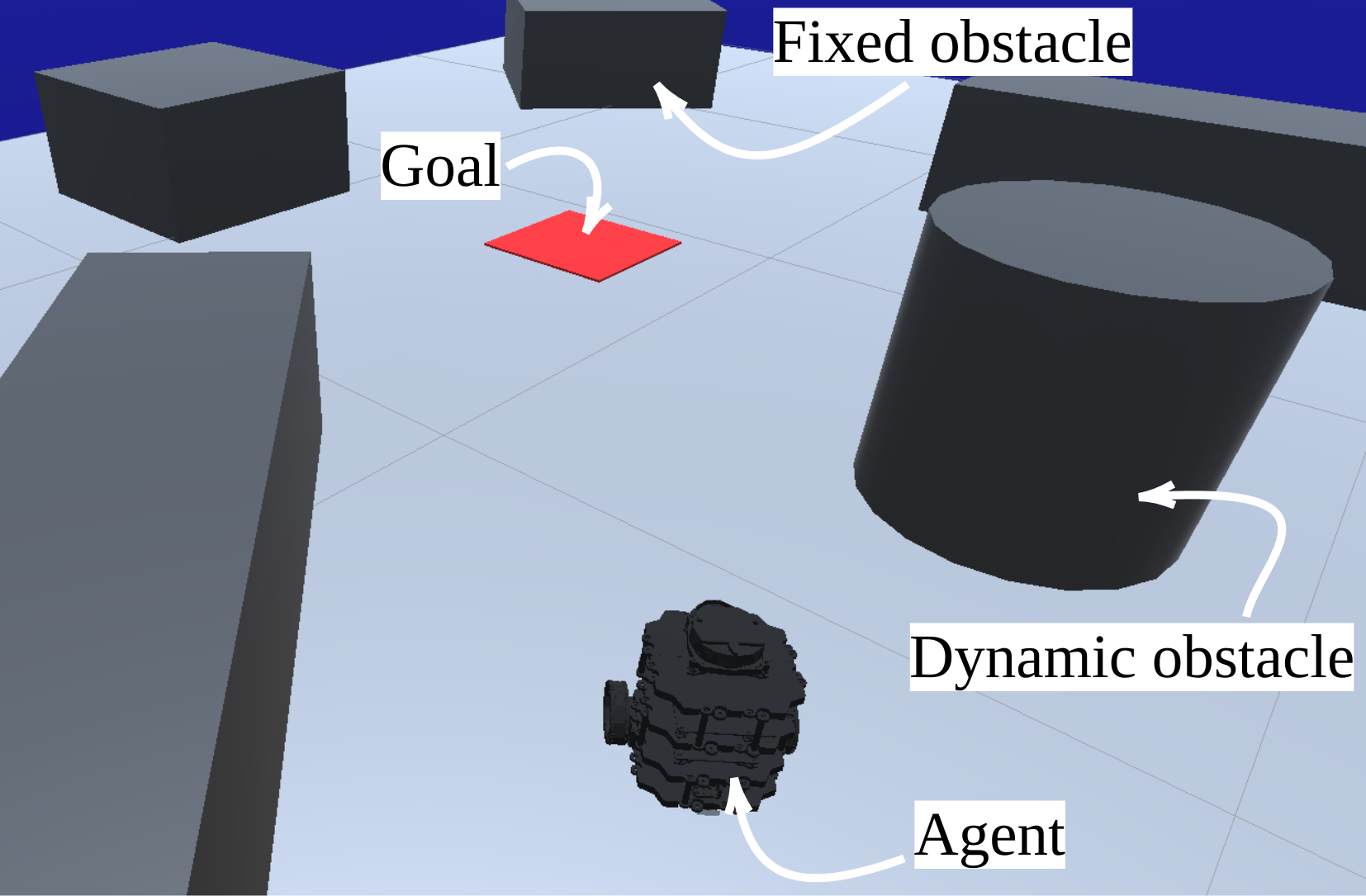}
        \caption{Overview of the \textit{Evaluation} environment}
        \label{fig:envs}
    \end{center}
    \vspace{-3mm}
\end{figure}

\begin{figure*}[t]	
    \begin{center}
    	\includegraphics[width=.9\linewidth]{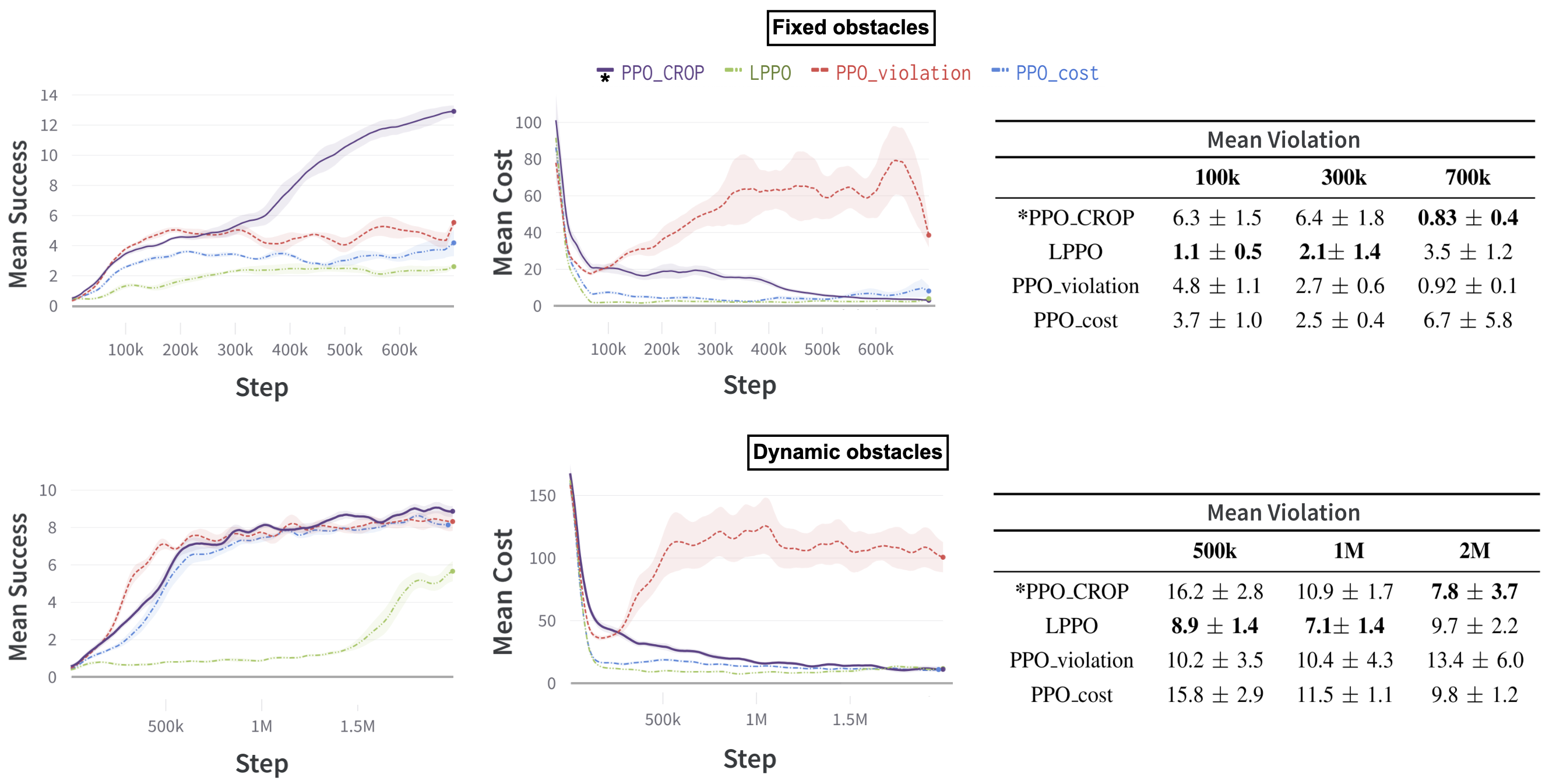}
        \caption{Average success, cost, and violation for our versions of PPO\_cost, PPO\_violation, \textbf{*}PPO\_CROP and LPPO. \textbf{*} indicates our method.}
        \label{fig:results}
    \end{center}
    \vspace{-7mm}
\end{figure*}

More specifically, the environments differ for obstacles that can be \textit{Fixed} (parallelepiped-shaped static objects) or \textit{Dynamic} (cylindrical-shaped objects that move to random positions with constant velocity). The environments have \textit{non-terminal} obstacles, i.e., they return a signal upon each collision, but the episode does not terminate (i.e., the robot has to learn how to avoid getting stuck into the obstacles). The algorithms use such collisions to trigger a positive value or other desired penalties. Dynamic obstacles are particularly challenging because they lead to a non-stationary environment and make the problem partially observable \cite{POMDP}. To this end, we extend the setup of previous work\cite{drl_navigation2, AAAI_Sos} to consider 360-degree scans. This setup allows us to obtain a good trade-off between mean success and obstacle avoidance behaviors. Finally, we test the generalization skills of the trained policies in an \textit{Evaluation} environment that is not employed during training, depicted in Fig. \ref{fig:envs}.

\subsection{Motivating example}
We conducted an experiment to confirm our hypotheses regarding the limitation of hard-coded properties-based approaches in the fixed obstacle domain. In particular, we compare the number of properties employed to compute the violation value (i.e., the properties that contains the unsafe interaction in their domain-codomain) in the case of PPO\_CROP (ours) and PPO\_violation (which considers a set of five hard-coded properties for navigation as in previous work \cite{prove}). In this experiment, we expect PPO\_CROP to collect a higher number of properties that describe the actual unsafe behaviors experienced during the training. Hence, the number of properties used to compute the violation in the case of PPO\_CROP should be significantly higher. As discussed in the following empirical evaluation, such a higher number of properties lead to a more informative \textit{violation} value (i.e., it is computed considering a higher number of unsafe situations), hence improving the safety aspect of the trained policies.
\begin{figure}[b]
   \vspace{-1mm}
    \begin{center}
    	\includegraphics[width=0.75\linewidth]{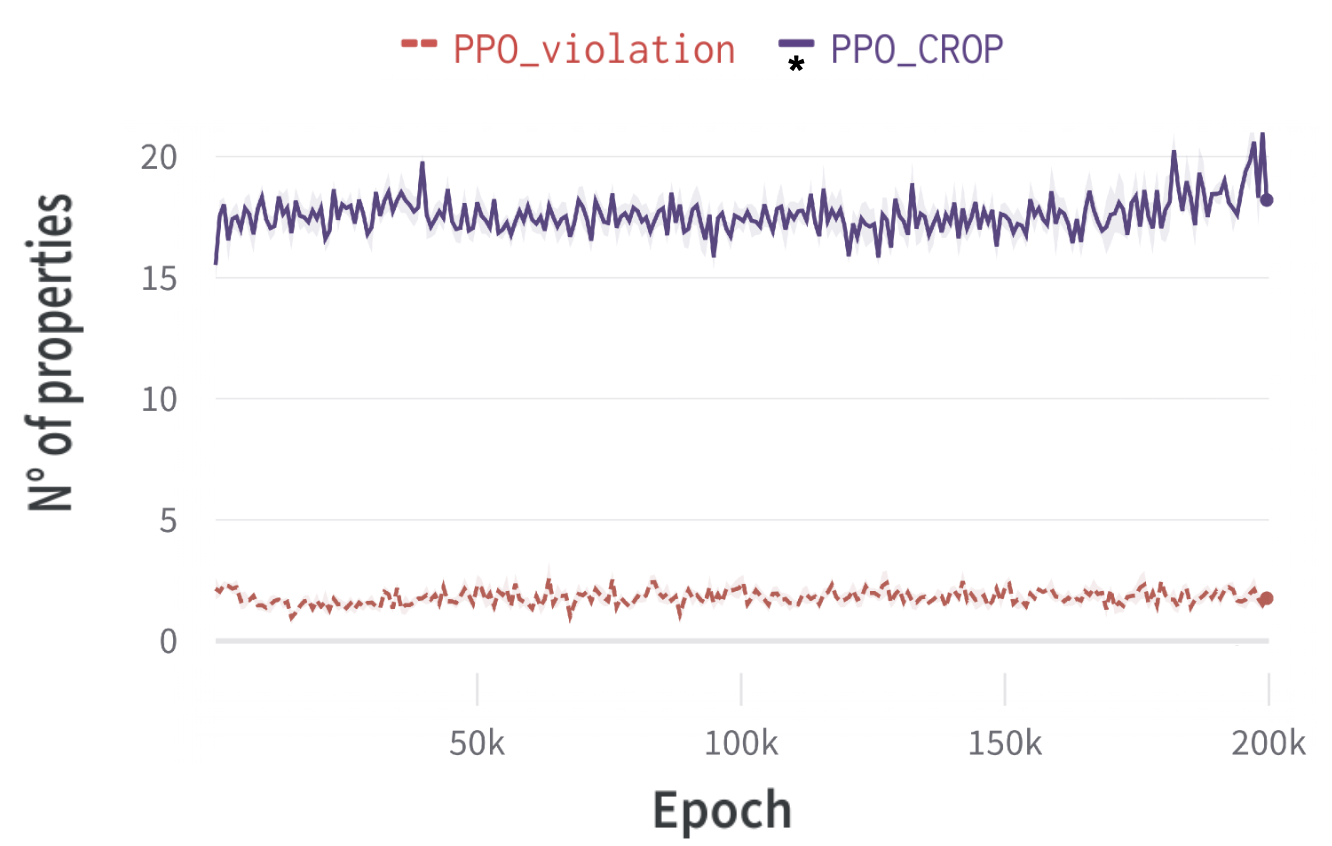}
        \caption{Average number of proprieties' used for the \textit{violation} computation during the training of PPO\_violation and our PPO\_CROP.}
        \label{fig:motivating}
    \end{center}
    \vspace{-4mm}
\end{figure}

Fig.\ref{fig:motivating} shows the number of properties used by PPO\_violation and PPO\_CROP to compute the \textit{violation}. As expected, CROP collects a high number of properties on the task, while prior work cannot cope with the complexity of the environment and employs a reduced number of properties for computing the \textit{violation}. 

Finally, we note that the number of properties used in CROP tends to increase by the end of the training, while it remains constant for the other methodology. Such result implies that collecting properties during the training by the agent allows it to find borderline cases of unsafe situations that are difficult to explore using hand-designed properties.
\begin{table*}[t]
    \caption{Mean violation expressed in percentage values computed using Formal Verification for each models at convergence for \textit{Dynamic obstacles} (left) and \textit{Fixed obstacles} (right) environments.}
    
    \label{tab:verification}
    \begin{minipage}{.48\linewidth}
       \begin{adjustbox}{max width=\textwidth}
        \begin{tabular}{c|cccc}
         \toprule
         & \multicolumn{4}{c}{\textbf{Method}} \\ \toprule
        \textbf{Property} & \textbf{PPO\_cost} & \textbf{*PPO\_CROP} & \textbf{PPO\_violation} & \textbf{LPPO} \\ \midrule \smallskip
        $\mathcal{P}_{\uparrow}$ & 0.2 $\pm$ 0.04 & 0.09 $\pm$ 0.07 & 0.15 $\pm$ 0.1 & 0.29 $\pm$ 0.1 \\\smallskip
        $\mathcal{P}_{\rightarrow}$ & 0.47 $\pm$ 0.04 & 0.45 $\pm$ 0.1 & 0.41 $\pm$ 0.2 & 0.38 $\pm$ 0.1 \\\smallskip
        $\mathcal{P}_{\leftarrow}$ & 0.51 $\pm$ 0.2 & 0.43 $\pm$ 0.08 & 0.39 $\pm$ 0.1 & 0.43 $\pm$ 0.3 \\\smallskip
        $\mathcal{P}_{\nearrow}$ & 0.49 $\pm$ 0.09 & 0.36 $\pm$ 0.08 & 0.46 $\pm$ 0.1 & 0.49 $\pm$ 0.3 \\\smallskip
         $\mathcal{P}_{\nwarrow}$& 0.5 $\pm$ 0.1 & 0.33 $\pm$ 0.07 & 0.42 $\pm$ 0.2 & 0.44 $\pm$ 0.2 \\ \midrule
        \textbf{SUM} & 2.17 & \textbf{1.66} & 1.83 & 2.03
        \end{tabular}
        \end{adjustbox}
    \end{minipage}%
    \quad
    \begin{minipage}{.48\linewidth}
      \centering
         \begin{adjustbox}{max width=\textwidth}
        \begin{tabular}{c|cccc}
        \toprule
         & \multicolumn{4}{c}{\textbf{Method}} \\ \toprule
        \textbf{Property} & \textbf{PPO\_cost} & \textbf{*PPO\_CROP} & \textbf{PPO\_violation} & \textbf{LPPO} \\ \midrule \smallskip
        $\mathcal{P}_{\uparrow}$ & 0.3 $\pm$ 0.1 & 0.26 $\pm$ 0.2 & 0.4 $\pm$ 0.2 & 0.28 $\pm$ 0.1 \\\smallskip
        $\mathcal{P}_{\rightarrow}$ & 0.52 $\pm$ 0.1 & 0.38 $\pm$ 0.1 & 0.35 $\pm$ 0.2 & 0.34 $\pm$ 0.3 \\\smallskip
        $\mathcal{P}_{\leftarrow}$ & 0.63 $\pm$ 0.3 & 0.42 $\pm$ 0.2 & 0.35 $\pm$ 0.1 & 0.63 $\pm$ 0.4 \\\smallskip
        $\mathcal{P}_{\nearrow}$ & 0.57 $\pm$ 0.2 & 0.35 $\pm$ 0.1 & 0.39 $\pm$ 0.2 & 0.57 $\pm$ 0.3 \\\smallskip
         $\mathcal{P}_{\nwarrow}$& 0.56 $\pm$ 0.2 & 0.33 $\pm$ 0.1 & 0.4 $\pm$ 0.2 & 0.4 $\pm$ 0.3 \\ \midrule
        \textbf{SUM} & 2.58 & \textbf{1.74} & 1.89 & 2.22
        \end{tabular}
        \end{adjustbox}
    \end{minipage} 
    \vspace{-3mm}
\end{table*}
\subsection{Empirical training evaluation}
Fig.\ref{fig:results} shows the results of our evaluation in \textit{Fixed obstacles} and \textit{Dynamic obstacles}. In more detail, in the \textit{Fixed obstacles} environment, our PPO\_CROP outperforms other methodologies in terms of the average success rate and significantly reduces the cost (i.e., number of collisions) and \textit{violation} during the training. In this environment, PPO\_CROP at convergence shows a $\approx 0.09\%$ and a $\approx 5.87\%$ \textit{violation} improvement over the PPO\_violation and PPO\_cost counterparts. Considering the total number of points used to compute the violation (which depends on the number of properties employed by the approaches), such improvements map to $180$ and $11740$ fewer collisions, respectively.
Regarding the \textit{Dynamic obstacles} environment, our methodology achieves better or comparable successes and cost values over the counterparts but significantly reduces the violations during the training, showing a $\approx 5.60\%$ and a $\approx 2.00\%$ \textit{violation} improvement, which corresponds on average to $11200$ and $4000$ fewer collisions. For a fair evaluation, all the algorithms are trained over the same parameters and configurations of dynamic obstacles (i.e., they experience the same random obstacle movements during the training phases, which changes over the different seeds).
Finally, LPPO satisfies the constraint imposed on the cost threshold, set respectively at $5$ and $12$ in the \textit{Fixed obstacles} and \textit{Dynamic obstacles} environments. However, it achieves the lowest number of successes, confirming the performance trade-off in complex scenarios and the difficulty of tuning the value of the multiplier parameter underlined in other recent research works \cite{SafetyGym, AAAI_Sos}. Crucially, the lower number of successes is not motivated by performing longer but safer paths, as PPO\_CROP also achieves better results in terms of safety (i.e., lower violations and cost).

\subsection{Evaluation}
In order to test the generalization skills of the trained policies in a previously unseen scenario, we perform an additional experiment on the \textit{Evaluation} environment (reported in Fig.\ref{fig:envs}). 
\begin{table}[h!]
    \centering
    \caption{Evaluation results in the \textit{Evaluation} environment}
    \vspace{-5mm}
    \label{tab:eval}
   \begin{adjustbox}{max width=\linewidth}
        \begin{tabular}{cccc}
        \multicolumn{1}{c}{} & \multicolumn{1}{c}{} \\ \toprule
        \textbf{Method} & \textbf{Mean Success} & \textbf{
        Mean Cost} & \textbf{Mean Violation} \\ \midrule   \smallskip
        PPO\_cost & 8.4 $\pm$ 1.5 & 2.4 $\pm$ 0.5 & 2.1 $\pm$ 0.5\\\smallskip
        \textbf{*}PPO\_CROP & \textbf{8.8} $\pm$ 0.8 & \textbf{1.8} $\pm$ 0.9 & \textbf{1.3} $\pm$ 0.6 \\\smallskip
        PPO\_violation & 8.2 $\pm$ 0.6 & 84.9 $\pm$ 28.6 & 79.7 $\pm$ 30.7 \\\smallskip
        LPPO & 4.0 $\pm$ 0.9 & 2.1 $\pm$ 1.6 & 1.8 $\pm$ 1.3 \\
        \end{tabular}
     \end{adjustbox}
\end{table}
Tab. \ref{tab:eval} reports each model's average success, cost, and \textit{violation} at convergence. Results confirm the benefit of using our CROP, as PPO\_CROP shows superior navigation skills by achieving a higher number of successes while being safer over LPPO and the cost counterpart.

\subsection{Formal Verification at convergence}
 We provide formal guarantees on the trained policies behaviors using recent FV literature \cite{prove} in reinforcement learning scenarios. In contrast to the sample-based estimation used during the training, formal verification framework provably ensures the policy decisions in the situations specified in safety properties (i.e., it does not depend on the samples, but it is an exhaustive search over the entire space described by the properties' domains), therefore it provides guarantees that the behavior of the trained model respects a set of safety properties. Moreover, such formal verification frameworks also return all the states where a policy violates the properties. For a fair evaluation of the policies at convergence, we verify all the policies using the five hard-coded properties of prior work\cite{AAAI_Sos}. 
 Tab. \ref{tab:verification} shows the average violations computed over such properties. 
 Crucially, the formal analysis of the trained networks confirms our empirical results. Overall, PPO\_CROP results the safest approach, with the lowest \textit{violation} rate in both the \textit{Dynamic} and \textit{Fixed} obstacles environment, which translates into fewer collisions.

\subsection{Real-Robot experiments}
\begin{figure}[h!]
    \centering
    \includegraphics[width=\linewidth]{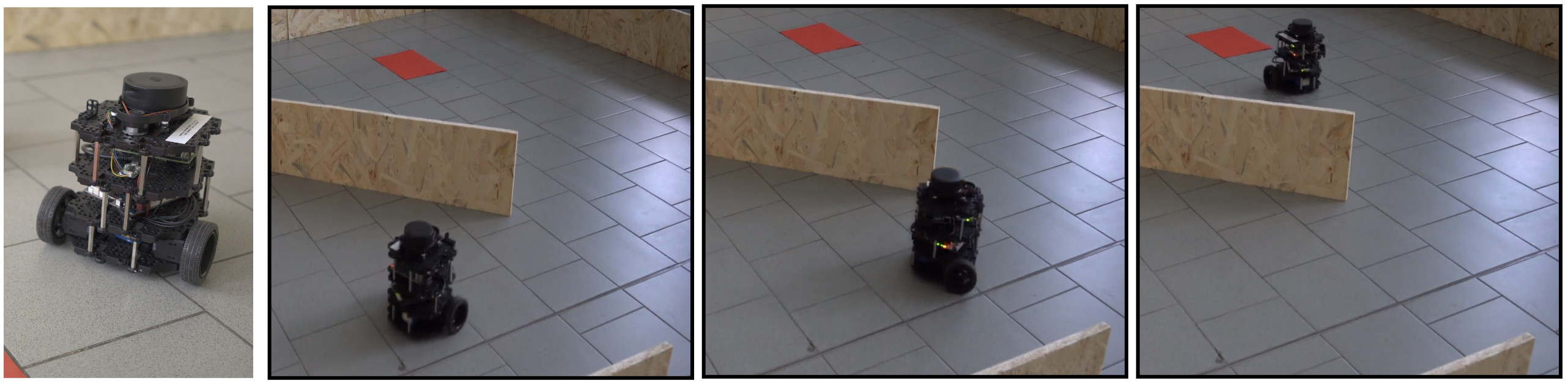}
    \caption{Real-world experiments.}
    \label{fig:real_exp}
    \vspace{-3mm}
\end{figure}

The Unity framework\cite{unity} allowed us to transfer the policies trained in simulation on ROS-enabled platforms such as our Turtlebot3 (Fig. \ref{fig:real_exp}). In particular, we show a comparison between our PPO\_CROP and PPO\_violation in several real corner case scenarios\footnote{Video \url{shorturl.at/drVW6} of real-world experiments.}.


\section{Discussion}
\label{sec:discussion}
This paper addresses the limitations of safety-oriented approaches that leverage only hard-coded properties. The proposed framework, CROP, allows the collection and refinement of safety properties during the training, thus overcoming the limitations of hand-designed approaches. We show that CROP allows us to obtain a more robust approach with respect to other Safe DRL methodologies, promoting safer behaviors while maintaining similar or better returns. 

Future directions involve extending the use of CROP where safety is of pivotal importance, such as multi-agent systems where the properties have to consider cooperative situations.

\clearpage

\bibliographystyle{IEEEtran}
\bibliography{biblio.bib}


\end{document}